\documentclass[conference]{IEEEtran}
\IEEEoverridecommandlockouts

\usepackage{cite}
\usepackage{amsmath,amssymb,amsfonts}
\usepackage{algorithmic}
\usepackage{graphicx}
\usepackage{textcomp}
\usepackage{xcolor}
\usepackage{booktabs}
\def\BibTeX{{\rm B\kern-.05em{\sc i\kern-.025em b}\kern-.08em
    T\kern-.1667em\lower.7ex\hbox{E}\kern-.125emX}}
\begin{document}

\title{BaT-SAM: Introducing Inductive Bias into SAM via a Mixture of Convolutional Experts for Brain Tumor Segmentation\\
}
\author{\IEEEauthorblockN{Sumiaya Jannat}
\IEEEauthorblockA{\textit{Mymensingh Medical College} \\
Mymensingh, Bangladesh \\
sumiaya.m55@gmail.com}
\and
\IEEEauthorblockN{Md Mahedi Hasan}
\IEEEauthorblockA{\textit{Department of LCSEE} \\
\textit{West Virginia University}\\
Morgantown, WV, USA \\
mh00062@mix.wvu.edu}
\and
\IEEEauthorblockN{Md Shamimul Islam}
\IEEEauthorblockA{\textit{Department of CS} \\
\textit{Florida Polytechnic University}\\
Lakeland, Florida, USA \\
mislam3051@floridapoly.edu}
\and
\IEEEauthorblockN{Md Nur Hossain}
\IEEEauthorblockA{\textit{Department of CSE} \\
\textit{Manarat International University}\\
Dhaka, Bangladesh \\
nurhossainjoy69@gmail.com}

 \thanks{This work has been submitted to the IEEE for possible publication. 
Copyright may be transferred without notice, after which this version may no longer be accessible.}

}

\maketitle
\begin{abstract}
Accurate brain tumor segmentation from magnetic resonance imaging (MRI) is essential for diagnosis and treatment planning; however, manual delineation is labor-intensive and time-consuming. Existing approaches are either task-specific with limited generalization or rely on user-provided prompts during inference. We present BaT-SAM, a parameter-efficient adaptation framework that enables fully automatic brain tumor segmentation with the Segment Anything Model (SAM). BaT-SAM injects local inductive bias into SAM's frozen Vision Transformer (ViT) encoder via Conv-LoRA, where image features are processed by a sparsely gated mixture of multi-scale convolutional experts. This design compensates for the limited translation-equivariant inductive bias of vanilla ViTs while fine-tuning fewer than 5M parameters ($<$1\% of the full model). To remove the need for user prompts, we further introduce a lightweight self-prompting box regressor that is jointly optimized with the adapter using scheduled sampling and predicts tumor locations directly from image embeddings. On the BRISC2025 benchmark, BaT-SAM achieves a weighted mean IoU of 0.832 while outperforming state-of-the-art benchmarks. It further demonstrates strong cross-dataset generalization, achieving Dice scores of 0.702 on Figshare and 0.784 on the LGG MRI Segmentation dataset.
\end{abstract}


\section{Introduction}
Accurate segmentation of brain tumors from MRI images, including gliomas, meningiomas, and pituitary adenomas, is critical for diagnosis, radiotherapy planning, and treatment monitoring. Despite recent advances, tumor delineation in clinical practice remains largely dependent on slice-by-slice manual annotation, a labor-intensive and time-consuming process. Automatic segmentation is particularly challenging due to substantial variability in tumor location, size, shape, and appearance across different tumor types. Moreover, tumor growth patterns differ considerably. For example, gliomas often exhibit diffuse, infiltrative boundaries that blend with surrounding edema, whereas meningiomas and pituitary adenomas are generally more localized but may still present irregular morphologies. These factors contribute to significant inter-annotator variability, making manual delineation increasingly impractical. Therefore, developing accurate and robust automatic segmentation methods that generalize across tumor types, imaging scanners, and patient populations remains an important clinical and research objective.

U-Net~\cite{ronneberger2015unet} and its variants~\cite{oktay2018attention, zhou2018unetpp, isensee2021nnunet} have been widely used for brain tumor segmentation. However, these models are typically trained from scratch using densely annotated pixel-level masks, whose acquisition is both labor-intensive and expensive. As a result, tumor segmentation datasets remain orders of magnitude smaller than general image benchmarks, causing models to overfit limited training data and generalize poorly to cross-domain settings. Furthermore, convolutional networks primarily rely on local receptive fields which limits their ability to capture long-range anatomical dependencies that are often crucial for accurately delineating tumor boundaries and subregions. A promising alternative is to adapt large-scale vision foundation models that have learned transferable segmentation priors. For example, the Segment Anything Model (SAM)~\cite{sam_2023}, pre-trained on the SA-1B dataset containing over one billion segmentation masks, learns rich representations of object boundaries and fine-grained structures. Therefore, rather than fully fine-tuning such foundation on limited medical data, it is more effective to efficiently adapt its pre-trained representations to the brain MRI domain.

Adapting SAM to brain tumor MRI segmentation presents several challenges. First, a substantial domain gap exists between SAM's pre-training on natural images and gray-scale brain MRI scans. Compared with natural images, T1-weighted MRI slices exhibit distinct intensity distributions, textures, and noise characteristics, while brain tumors often appear as low-contrast, diffuse lesions rather than the well-defined, high-contrast objects on which SAM was trained. As a result, zero-shot SAM generalizes poorly to brain tumor MRI, achieving average Dice scores of only 0.56--0.70 depending on the prompt type~\cite{ali2024evaluating}. Second, fully fine-tuning SAM's ViT encoder on limited annotated data is prone to overfitting~\cite{xct_sam}, encouraging us to employ the parameter-efficient adaptation technique. Third, SAM is inherently designed for interactive segmentation that requires users to provide prompt in the form of points, bounding boxes, or masks during inference~\cite{zhang2026brainsam}. This requirement is incompatible with fully automatic clinical workflows which motivates us to develop an automatic prompt generation strategy for SAM~\cite{sam_2023}.

To address these challenges, we propose \textbf{BaT-SAM}, a robust, fully automatic, and parameter-efficient framework for \textbf{br}ain \textbf{t}umor segmentation based on \textbf{SAM}. We adapt SAM by injecting Conv-LoRA modules~\cite{zhong2024convlora} into its frozen ViT image encoder. Specifically, Conv-LoRA extends LoRA~\cite{hu2021lora} by routing low-rank features through a sparsely gated mixture of multi-scale convolutional experts before projecting them back into the frozen backbone. By incorporating convolutional inductive biases into the transformer encoder, we compensate for the lack of translation equivariance in the ViT encoder and improve the representations of fine-grained tumor boundaries. Our approach achieves these benefits while updating fewer than \textbf{1}\% of SAM's parameters.

Within the BaT-SAM framework, we employ a single shared adapter and segmentation decoder to jointly segment gliomas, meningiomas, and pituitary adenomas. To enable fully automatic inference, we introduce a lightweight self-prompting box regressor that is trained jointly with the segmentation network. The regressor predicts tumor bounding boxes directly from SAM's image embeddings and passes them to the prompt encoder, removing the need for manual prompts during inference. To further improve robustness, we adopt a scheduled sampling strategy that gradually replaces ground-truth box prompts with predicted boxes throughout training. This curriculum mitigates \textit{exposure bias} (train--test prompt discrepancy), resulting in more reliable fully automatic segmentation during inference.

We evaluate BaT-SAM on the BRISC2025 benchmark~\cite{fateh2025brisc}. Despite updating fewer than 5M parameters, BaT-SAM achieves \textbf{0.832} mIoU and \textbf{0.887} Dice, outperforming the previous state-of-the-art Swin-HAFUNet~\cite{fateh2025brisc}, which reports a 0.824 weighted mIoU. We further evaluate BaT-SAM on two independent external datasets to assess its generalization ability, achieving Dice scores of \textbf{0.702} on the Figshare Brain Tumor dataset~\cite{cheng2015enhanced} and \textbf{0.784} on the LGG MRI Segmentation dataset~\cite{buda2019lgg}. In summary, we make the following contributions:
\begin{itemize}
	\item We propose a parameter-efficient adaptation framework, BaT-SAM, that injects local, translation-equivariant inductive biases into SAM's frozen image encoder through a sparsely gated mixture of multi-scale convolutional experts, while updating fewer than 5M parameters.

	\item We introduce a self-prompting box regressor that is trained end-to-end with a scheduled sampling strategy, removing the need for manually provided prompts and enabling a single shared adapter and decoder to segment all brain tumor types.
	
	\item We demonstrate that BaT-SAM achieves state-of-the-art performance on the BRISC2025 benchmark and generalizes effectively to two independent brain MRI datasets in a zero-shot setting.
\end{itemize}

\section{Related Work}
\noindent\textbf{Brain tumor segmentation.} Convolutional encoder--decoder networks have long been the dominant paradigm for brain tumor segmentation. U-Net~\cite{ronneberger2015unet} established the standard encoder--decoder architecture with skip connections, while subsequent variants improved feature fusion through dense skip connections (U-Net++~\cite{zhou2018unetpp}) and attention mechanisms (Attention U-Net~\cite{oktay2018attention}). nnU-Net~\cite{isensee2021nnunet} further demonstrated that careful pipeline optimization often outperforms architectural modifications across biomedical segmentation benchmarks, including BraTS~\cite{bakas2018brats}. However, these methods rely on task-specific supervision, requiring extensive pixel-level annotations while demonstrate limited generalization to domain shifts caused by variations in scanners and imaging protocols.

\noindent\textbf{Adapting SAM for medical imaging.} Although the Segment Anything Model (SAM)~\cite{sam_2023} learns strong promptable segmentation priors from natural images, its performance degrades substantially on medical images because of the large domain gap with MRI or CT imaging domain. MedSAM~\cite{ma2024medsam} bridges this gap by fine-tuning SAM on a large-scale, multi-modal medical image dataset with corresponding segmentation masks, achieving significant improvements over zero-shot settings. In parallel, SAM-Med2D~\cite{cheng2023sammed2d} adapts SAM to 2D medical images by fine-tuning the prompt encoder and mask decoder while introducing lightweight adapters into the image encoder, showing strong generalization across a broad range of medical imaging tasks.

\noindent\textbf{SAM for brain tumor segmentation.} Ali \textit{et al.}~\cite{ali2024evaluating} evaluated SAM in a zero-shot setting on two brain tumor MRI datasets and reported lower Dice scores of 0.56--0.70, depending on the dataset. They further showed that bounding box prompts consistently outperform point prompts for irregular tumor segmentation. Building on this observation, Ryu \textit{et al.}~\cite{ryu2025enhancing} employed a vision--language model to generate box prompts before fine-tuning the mask decoder. However, the predicted boxes remain noisy, achieving a mean box IoU of only 0.23--0.36, which limits segmentation performance. More recently, EffSAM~\cite{ahmad2025effsam} removed the prompt encoder entirely and adapted SAM for volumetric brain MRI using low-rank adaptation, and cross-slice attention. In contrast, we preserve SAM's prompting paradigm by learning a lightweight self-prompting box regressor that enables fully automatic inference while efficiently adapting the frozen image encoder.

\noindent\textbf{Parameter-efficient adaptation.} Low rank adaptation (LoRA)~\cite{hu2021lora} adapts pre-trained models by learning low-rank updates to selected weight matrices, substantially reducing the number of trainable parameters compared with full fine-tuning. Conv-LoRA~\cite{zhong2024convlora} extends LoRA by routing low-rank features through a mixture of multi-scale convolutional experts, injecting inductive biases into SAM. This strategy has recently shown promising results for adapting SAM to industrial X-ray CT defect segmentation~\cite{tabassum2024adapting, xct_sam}. In this work, we build upon Conv-LoRA and extend it to brain tumor MRI segmentation.

\section{Method}
\label{sec:method}

\subsection{Overview}
\begin{figure*}[t]
	\centering
	\includegraphics[width=0.85\textwidth]{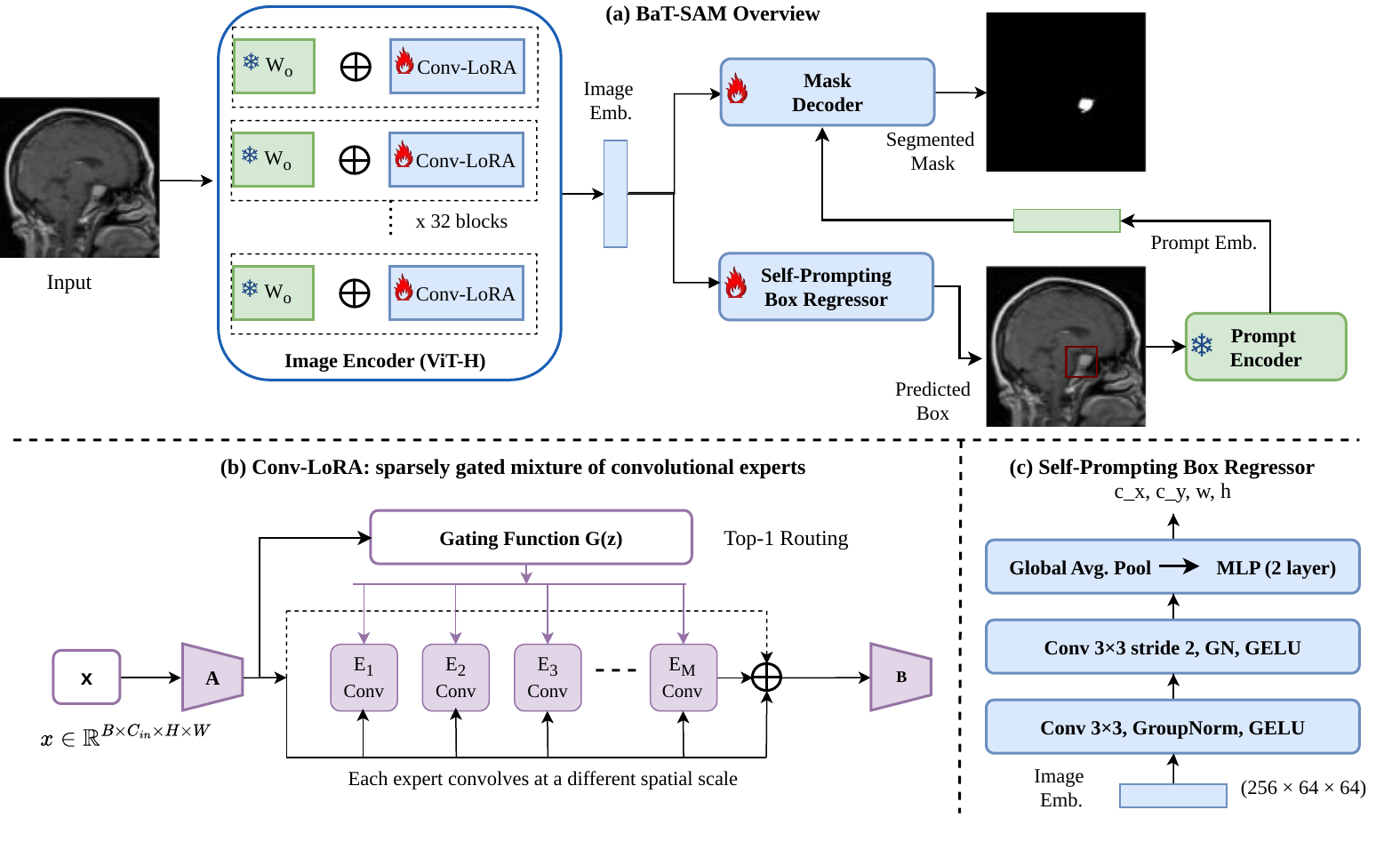}
	\vspace{-6mm}
	\caption{Overview of BaT-SAM. (a) Conv-LoRA adapters are inserted into the frozen SAM ViT-H encoder. (b) Each Conv-LoRA adapter routes the rank-$r$ bottleneck through a sparsely gated mixture of multi-scale convolutional experts, injecting local inductive bias absent in standard LoRA. (c) A self-prompting box regressor, consisting of a convolutional stem, global average pooling, and a lightweight MLP, predicts the bounding-box prompt for the prompt encoder.}
	\label{fig:architecture}
\end{figure*}
Figure~\ref{fig:architecture} provides an overview of the proposed BaT-SAM framework. BaT-SAM builds upon the ViT-H variant of SAM, which consists of an image encoder, a prompt encoder, and a mask decoder. To efficiently adapt SAM to brain MRI segmentation, we inject Conv-LoRA modules into the frozen ViT image encoder, introducing local translation-equivariant inductive biases. To enable fully automatic inference, we further introduce a lightweight box regressor that predicts tumor bounding boxes directly from the image embeddings and feeds them to the prompt encoder. We jointly train the box regressor and segmentation network using a scheduled sampling strategy.

\subsection{Conv-LoRA for Inductive Bias}
As SAM's image encoder is a standard vision transformer, it lacks the local inductive biases inherent to convolutional networks. To address this limitation, we inject these inductive biases using Conv-LoRA. Specifically, we apply Conv-LoRA only to the combined query, key, and value (QKV) projection in each of the 32 transformer blocks of the image encoder. Given a frozen projection matrix $W_0$, the trainable adaptation is defined as
\begin{equation}
h = W_0 x + \frac{\alpha}{r} B A x,
\label{eq:lora}
\end{equation}

where $A \in \mathbb{R}^{r \times d}$ and $B \in \mathbb{R}^{d \times r}$ are learnable low-rank matrices, $r$ denotes the rank, and $\alpha$ is a scaling factor. Following~\cite{hu2021lora}, we initialize $B$ to zero so that the adapted model is identical to the frozen SAM. Conv-LoRA extends LoRA by transforming the low-rank bottleneck representation, $z = Ax$, with a sparsely gated mixture of $M$ multi-scale convolutional experts. Specifically, we reshape $z$ into a spatial feature map and route it through experts with different effective receptive fields, obtained by lightweight upsampling and downsampling operations. A gating network predicts expert weights from globally pooled bottleneck features and dynamically selects the most suitable expert for each input. We use rank $r=2$ and $M=8$ experts. We add an auxiliary load balancing loss on the gating network, to discourage the gate from collapsing onto a small subset of experts.

\subsection{Self-Prompting Box Regressor}
SAM requires user-provided prompts during inference. We remove this dependency by introducing a lightweight self-prompting box regressor that predicts tumor bounding boxes directly from the 256-channel image embedding using two convolutional blocks, global average pooling, and a two-layer MLP. We regress the box center and size instead of the box corners for more stable optimization.
\begin{equation}
(c_x, c_y, w, h) = \sigma\big(\text{MLP}(\text{GAP}(\text{Conv}(E)))\big).
\label{eq:boxreg}
\end{equation}
Here $E$ denotes the image embedding and $\sigma$ is the sigmoid function. We convert the predicted box center and size into corner coordinates, clamp them to $[0,1]$, and scale the resulting box to pixel coordinates before passing it to the frozen prompt encoder. This parameterization ensures a valid bounding box with non-negative width and height.

\subsection{Expanded Ground Truth Box}
Some tumors, particularly gliomas, consists of multiple disjoint lesions that are not fully enclosed by a tight bounding box. To improve coverage, in this study, we enlarge the ground-truth box by a fixed margin before training:
\begin{equation}
	\hat{b}_{\text{gt}} = \mathrm{expand}(b_{\text{gt}}, \gamma),
	\label{eq:padbox}
\end{equation}
where $b_{\text{gt}}$ denotes the tight ground-truth box, $\gamma$ is the expansion ratio. We set $\gamma=0.15$ in all experiments. The expanded box $\hat{b}_{\text{gt}}$ is used both as the regression target and as the reference box for the scheduled sampling curriculum described next. This encourages the regressor to predict boxes with sufficient lesion coverage while aligning the decoder's training and inference box distributions.

\subsection{Scheduled Sampling Curriculum}
At inference, the decoder receives only the regressor's predicted box, creating a train--test mismatch if trained solely with ground-truth boxes. To close this gap, in this study, we adopt a scheduled sampling curriculum scheme. Early in training, the decoder is conditioned on a jittered version of the expanded ground-truth box $\hat{b}_{\text{gt}}$, so the segmentation loss can still guide the decoder while the regressor is still learning. We jitter this box by applying small random translations and scale perturbations. As training progresses, we gradually anneal the probability of using $\hat{b}_{\text{gt}}$ to zero,
\begin{equation}
p(t) = \max\!\left(0,\; 1 - \frac{t}{\beta T}\right),
\label{eq:teacherforcing}
\end{equation}
where $t$ is the current epoch, $T$ is the total number of epochs, and $\beta$ controls the annealing duration. We set $\beta=0.8$, providing additional supervision for tumor classes with greater spatial variability, such as glioma. For each training sample, we draw a Bernoulli variable with probability $p(t)$ to determine whether the decoder receives the jittered ground-truth box or the regressor's prediction. By the end of training, the decoder is conditioned exclusively on predicted boxes, matching inference. The box regressor is always supervised against the expanded ground-truth box $\hat{b}_{\text{gt}}$, regardless of the box used to prompt the decoder.


\begin{table}[t]
	\centering
	\footnotesize 
	\setlength{\tabcolsep}{3pt}
	\caption{Statistics of the BRISC2025 segmentation split by tumor class. Location and shape statistics are computed over all annotated masks; location normalized centroid ($x,y \in [0,1]$).}
	\label{tab:brisc_seg_stats}
	\begin{tabular}{lcccc}
		\toprule
		Class &Train / Test &Area (\%) & \textbf{$x$ ($\mu \pm \sigma$)} & \textbf{$y$ ($\mu \pm \sigma$)}\\
		\midrule
		Glioma      & 1147 / 254 & $1.87 \pm 1.48$ & $0.49 \pm 0.11$ & $0.43 \pm 0.11$ \\
		Meningioma  & 1329 / 306 & $2.25 \pm 1.83$ & $0.50 \pm 0.13$ & $0.44 \pm 0.15$ \\
		Pituitary   & 1457 / 300 & $1.04 \pm 0.73$ & $0.50 \pm 0.05$ & $0.51 \pm 0.08$\\
		\midrule
		\textbf{Total} & \textbf{3933 / 860} & -- & -- & --\\
		\bottomrule
		\vspace*{-5mm}
	\end{tabular}
\end{table} 
\subsection{Training Objective}
We train BaT-SAM with three loss terms. The segmentation loss is a weighted mean of a Dice loss and a focal loss between the predicted mask and the ground truth mask,
\begin{equation}
\mathcal{L}_{\text{seg}} = \text{mean}_i\big[w_i\,(\mathcal{L}_{\text{Dice},i} + \mathcal{L}_{\text{Focal},i})\big].
\label{eq:segloss}
\end{equation}
The box loss is likewise a weighted mean of an $\ell_1$ loss and a generalized intersection over union (GIoU) loss between the regressor's predicted box and the padded ground truth box $\hat{b}_{\text{gt}}$ from Eq.~\ref{eq:padbox}, in normalized image coordinates,
\begin{equation}
\mathcal{L}_{\text{box}} = \text{mean}_i\Big[w_i\Big(\lVert b_{\text{pr},i} - \hat{b}_{\text{gt},i} \rVert_1 + 1 - \text{GIoU}(b_{\text{pr},i}, \hat{b}_{\text{gt},i})\Big)\Big].
\label{eq:boxloss}
\end{equation}
We also add the load balancing loss $\mathcal{L}_{\text{moe}}$ from every Conv-LoRA layer in the encoder. Our final training objective is
\begin{equation}
\mathcal{L} = \mathcal{L}_{\text{seg}} + \lambda_{\text{box}} \mathcal{L}_{\text{box}} + \lambda_{\text{moe}} \mathcal{L}_{\text{moe}},
\label{eq:totalloss}
\end{equation}
where we set $\lambda_{\text{box}} = \lambda_{\text{moe}} = 1$ in all experiments.

\begin{table*}[t]
	\centering
	\footnotesize
	\caption{Comparison of BaT-SAM on BRISC2025, LGG MRI, and Figshare brain tumor dataset for in-domain and cross-domain evaluation.}
	\label{tab:cross_dataset}
	\begin{tabular}{llllccc}
		\toprule
		Generalization &Evaluation &Method &Base &Type &Dice &IoU \\
		\midrule
		in-domain &Fine-tuning &Swin-HAFUNet~\cite{fateh2025brisc} & UNet variant & BRISC2025 $\to$ BRISC2025 & 0.8624 & 0.8240 \\
		in-domain &Fine-tuning &BaT-SAM (ours) & SAM &BRISC2025 $\to$ BRISC2025 & \textbf{0.8865} & \textbf{0.8325} \\
		\midrule
		in-domain &Fine-tuning &Attention U-Net~\cite{visalakshi2025sdhl} & UNet variant & LGG $\to$ LGG  & \textbf{0.9016} & 0.8078 \\
		in-domain &Fine-tuning &Res-UNet~\cite{huang2024segmenting} & UNet variant & LGG $\to$ LGG  & 0.8530 & 0.7590 \\
		in-domain &Fine-tuning &BaT-SAM (ours) &SAM & LGG $\to$ LGG  & 0.8561 & 0.\textbf{8090} \\
		
		\midrule 
		in-domain &Fine-tuning &Residual-UNet~\cite{alshomrani2024residualunet} & UNet variant & Figshare $\to$ Figshare & 0.7369 & 0.6151 \\
		in-domain &Fine-tuning &BaT-SAM (ours) & SAM &Figshare $\to$ Figshare & \textbf{0.8619} & \textbf{0.8150} \\
		
		\midrule
		cross-domain &Zero-shot &SAM-Med2D~\cite{cheng2023sammed2d} & SAM & BRISC2025 $\to$ LGG &0.7532 &0.6724 \\
		cross-domain &Zero-shot &SAM-Med2D~\cite{cheng2023sammed2d} & SAM & BRISC2025 $\to$ Figshare &0.6612 &0.6031 \\
		cross-domain &Zero-shot &BaT-SAM (ours) & SAM & BRISC2025 $\to$ LGG & \textbf{0.7842} & \textbf{0.6963} \\
		cross-domain &Zero-shot &BaT-SAM (ours) & SAM & BRISC2025 $\to$ Figshare & \textbf{0.7021} & \textbf{0.6173} \\
		\bottomrule
		\vspace*{-3mm}
	\end{tabular}
\end{table*}

\section{Experimental Setup}
\noindent\textbf{Datasets.} We train and evaluate BaT-SAM on the BRISC2025 benchmark~\cite{fateh2025brisc}, which contains 3,933 training and 860 test T1-weighted brain MRI slices ($512\times512$) with binary tumor masks. The dataset covers three tumor types (glioma, meningioma, and pituitary) across the axial, coronal, and sagittal planes, with one annotated tumor per slice and a balanced class distribution. As shown in Table~\ref{tab:brisc_seg_stats}, tumors occupy only 1.0--2.3\% of each slice on average, placing BRICS2025 MRI scans in the small-object segmentation regime. To evaluate cross-dataset generalization, we additionally test BaT-SAM on Figshare~\cite{cheng2015enhanced} and the LGG MRI Segmentation dataset~\cite{buda2019lgg}. The LGG dataset contains FLAIR MRI slices from 110 patients with lower-grade glioma and corresponding binary tumor masks. Since the annotations are defined on the FLAIR sequence, we use only the FLAIR images and replicate the single channel three times to match the BRISC2025 input format. We restrict evaluation to the 1,373 tumor-containing slices, as BaT-SAM assumes a single tumor per image.  

\begin{figure*}[t]
	\centering
	\includegraphics[width=\textwidth]{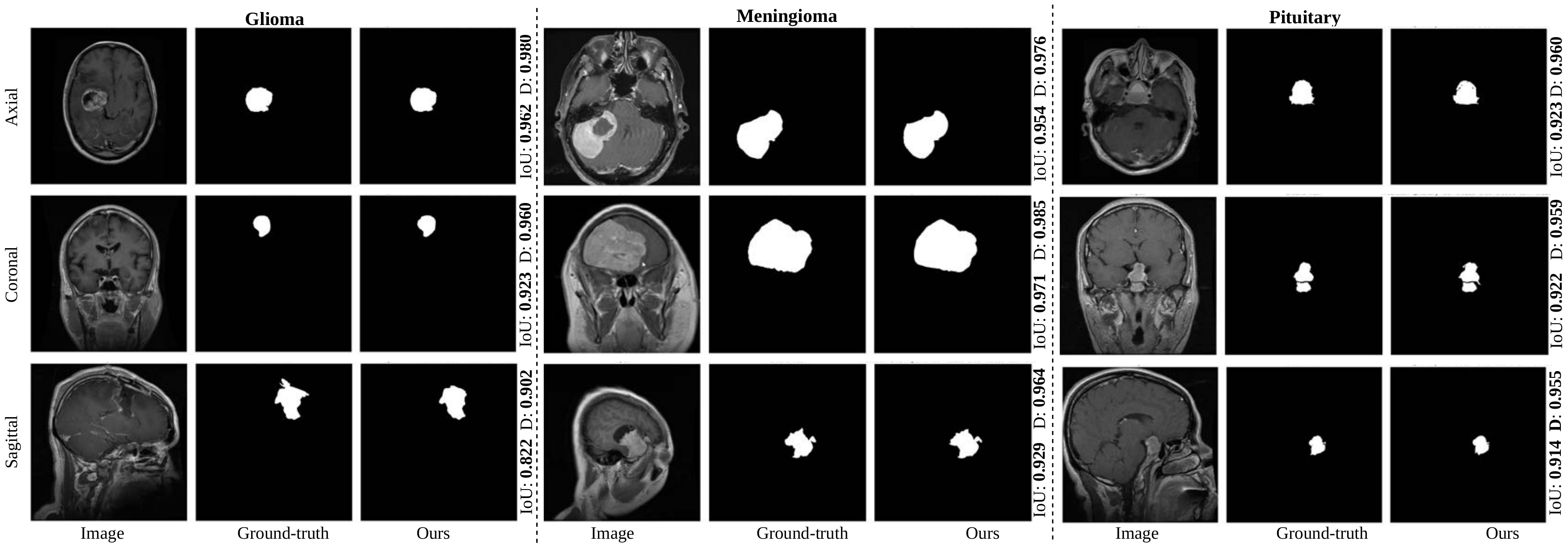}
	\caption{Qualitative segmentation results of BaT-SAM on the BRISC2025 test set. One representative example is shown for each tumor type across the axial, coronal, and sagittal planes. Each triplet consists of the input MRI slice, the ground-truth mask, and the corresponding BaT-SAM prediction.} 
	\label{fig:visualization}
	\vspace*{-3mm}
\end{figure*}

\noindent\textbf{Implementation Details.} We resize all images to $1024\times1024$, the native input resolution of SAM ViT-H. We also normalize them using ImageNet statistics and apply random horizontal flipping during training. We freeze the SAM image encoder and prompt encoder, training only the Conv-LoRA parameters, mask decoder, and box regressor, resulting in fewer than 5M trainable parameters ($<1\%$ of SAM ViT-H's 642M parameters). We optimize the model with AdamW, using a base learning rate of $3\times10^{-4}$ for the Conv-LoRA and mask decoder parameters and $3\times10^{-3}$ for the box regressor. We adopt a cosine learning rate schedule with linear warmup over the first 10\% of training and clip gradients to a maximum norm of 1. Training is performed with bfloat16 mixed precision and gradient accumulation, as the global attention blocks in SAM ViT-H are memory-intensive at full resolution. We fine-tune the model using distributed data parallelism on two RTX 6000 GPUs.

\noindent\textbf{Evaluation Metrics.} In this work, we report two metrics for mask quality. Intersection over union (IoU) measures the overlap between the predicted mask and the ground truth mask. In addition, the Dice score weights overlap relative to the combined size of both masks,
\begin{equation}
	\text{Dice} = \frac{2\,|m_{\text{pred}} \cap m_{\text{gt}}|}{|m_{\text{pred}}| + |m_{\text{gt}}|}.
	\label{eq:dice}
\end{equation}
Here, $m_{\text{pred}}$ is the predicted segmentation mask $m_{\text{gt}}$ is the ground truth mask. Unless otherwise stated, we report metrics averaged over all test samples, equivalent to weighting each class by its sample count, consistent with the BRISC2025 evaluation protocol~\cite{fateh2025brisc}.

\begin{table}[t]
	\centering
	\footnotesize
	\setlength{\tabcolsep}{3pt}
	\caption{IoU for brain tumor segmentation models on BRISC2025, by tumor type. mIoU is the sample-count-weighted average across tumor types.}
	\label{tab:brisc_baselines}
	\begin{tabular}{lcccc}
		\toprule
		Model &Glioma &Meningioma &Pituitary &mIoU \\
		\midrule
		UNet~\cite{ronneberger2015unet} & 0.697 & 0.771 & 0.793 & 0.757 \\
		UNet++~\cite{zhou2018unetpp} & 0.717 & 0.742 & 0.797 & 0.753 \\
		MANet~\cite{fan2020manet} & 0.724 & 0.775 & 0.780 & 0.762 \\
		LinkNet~\cite{chaurasia2017linknet} & 0.717 & 0.748 & 0.790 & 0.753 \\
		DeepLabV3+~\cite{chen2018deeplabv3plus} & 0.720 & 0.775 & 0.787 & 0.763 \\
		PAN~\cite{li2018pan} & 0.720 & 0.745 & 0.807 & 0.759 \\
		EINet~\cite{li2022einet} & 0.736 & 0.784 & 0.803 & 0.777 \\
		EU-Net~\cite{patel2021eunet} & 0.717 & 0.761 & 0.783 & 0.756 \\
		DAD~\cite{li2022dad} & 0.752 & 0.804 & 0.823 & 0.795 \\
		BASNet~\cite{qin2021basnet} & 0.740 & 0.775 & 0.817 & 0.779 \\
		SaberNet~\cite{saber2024sabernet} & 0.740 & 0.824 & 0.843 & 0.806 \\
		ABANet~\cite{rezvani2024abanet} & 0.724 & 0.804 & 0.847 & 0.795 \\
		Swin-HAFUNet~\cite{fateh2025brisc} & 0.760 & 0.850 & 0.853 & 0.824 \\
		\midrule
		\textbf{BaT-SAM (ours)} & \textbf{0.757} & \textbf{0.869} & \textbf{0.857} & \textbf{0.832} \\
		\bottomrule
	\end{tabular}
\end{table}

\section{Experimental Results}
\subsection{Main Results}
We first compare BaT-SAM with the BRISC2025 benchmark methods (Table~\ref{tab:brisc_baselines}), including convolutional and hybrid encoder--decoder architectures such as UNet~\cite{ronneberger2015unet}, UNet++~\cite{zhou2018unetpp}, and several recent variants. Among these, the previous state of the art is Swin-HAFUNet~\cite{fateh2025brisc}, a fully supervised Swin Transformer-based encoder--decoder augmented with hierarchical attention fusion and bottleneck enhancement modules. Unlike BaT-SAM, Swin-HAFUNet is trained by fully fine-tuning a task-specific segmentation network. It reports 82.4 percent weighted mIoU on the same test split, with 76.0 percent on glioma, 85.0 percent on meningioma, and 85.3 percent on pituitary. Despite updating only 5M parameters, BaT-SAM outperforms Swin-HAFUNet on BRISC2025, improving mean Dice from 0.8624 to 0.8865 (+2.4) and mean IoU from 0.8240 to 0.8325 (+0.9). On the LGG and Figshare datasets as reported in Table~\ref{tab:cross_dataset}, BaT-SAM also achieves the highest in-domain performance (mean Dice: 0.8561 and 0.8619) and demonstrates strong cross-domain generalization from BRISC2025, outperforming the SAM-Med2D baseline by 3.1\% and 4.0\% in Dice on LGG and Figshare, respectively.

\subsection{Visualization}
Figure~\ref{fig:visualization} presents qualitative results of BaT-SAM across all three tumor types and imaging planes in BRISC2025. The predicted masks closely align with the ground truth across diverse tumor morphologies, including irregular and elongated gliomas, where IoU remains above 0.8. Meningiomas, which are typically larger and more peripheral, achieve the highest overlap (IoU $> 0.95$, Dice $> 0.97$) in the axial and coronal views. Despite their small size, pituitary tumors are also segmented accurately across all three planes. These qualitative results validate our proposed framework. 

\section{Conclusion}\vspace*{-1mm}
We presented BaT-SAM, a parameter-efficient adaptation framework for brain tumor MRI segmentation based on the SAM. BaT-SAM augments the frozen ViT encoder with Conv-LoRA, which injects local inductive bias through a sparsely gated mixture of multi-scale convolutional experts while updating fewer than 5M parameters ($<1\%$ of the full model). A self-prompting box regressor, jointly optimized with the segmentation model via scheduled sampling, eliminates the need for user-provided prompts at inference. A single shared adapter and decoder generalize across all three tumor types without class-specific models. On the BRISC2025 benchmark, BaT-SAM achieves a weighted mIoU of 0.832, outperforming state-of-the-art methods. It further achieves higher Dice scores on Figshare and LGG MRI Segmentation dataset, demonstrating strong zero-shot cross-dataset generalization. In the future, we will investigate more robust localization strategies and broader training data diversity to further improve cross-domain generalization, especially for small and multifocal tumors.

\bibliographystyle{IEEEtranS.bst}
\bibliography{IEEEabrv.bib}
\end{document}